\documentclass{article}

\PassOptionsToPackage{numbers, sort&compress}{natbib}
\usepackage[final]{nips_2016}

\usepackage[utf8]{inputenc} 
\usepackage[T1]{fontenc}    
\usepackage{hyperref}       
\usepackage{url}            
\usepackage{booktabs}       
\usepackage{amsfonts}       
\usepackage{nicefrac}       
\usepackage{microtype}      
\usepackage{graphicx}

\newcommand{\subf}[2]{%
  {\small\begin{tabular}[t]{@{}c@{}}
  #1\\#2
  \end{tabular}}%
}
\graphicspath{ {C:\Users\Andy\Documents\MSc Thesis\Generative-and-Discriminative-Voxel-Modelling\paper} }
\title{Generative and Discriminative Voxel Modeling with Convolutional Neural Networks}

\author{
  Andrew Brock,
  Theodore Lim,
  J.M. Ritchie\\
  School of Engineering and Physical Sciences\\
  Heriot-Watt University\\
  Edinburgh, UK\\
  \texttt{\{ajb5, t.lim, j.m.ritchie\}@hw.ac.uk}
  \And Nick Weston\\
  Renishaw plc\\
  Research Ave, North\\
  Edinburgh, UK\\
  \texttt{Nick.Weston@renishaw.com}
}
\begin{document}
\maketitle
\begin{abstract}
When working with three-dimensional data, choice of representation is key. We explore voxel-based models, and present evidence for the viability of voxellated representations in applications including shape modeling and object classification. Our key contributions are methods for training voxel-based variational autoencoders, a user interface for exploring the latent space learned by the autoencoder, and a deep convolutional neural network architecture for object classification. We address challenges unique to voxel-based representations, and empirically evaluate our models on the ModelNet benchmark, where we demonstrate a 51.5\% relative improvement in the state of the art for object classification.
\end{abstract}

\section{Introduction}
\label{INTRO}
3D data offer computer vision systems a rich view of the world, but also pose a unique set of challenges, particularly in applications where understanding the surrounding environment is critical. In particular, data such as a point cloud extracted from an RGB-D image or a polygonal mesh are not guaranteed to be arranged in a regular grid, making them unsuitable for use with high-performance machine learning algorithms such as Convolutional Neural Networks (ConvNets). Deep ConvNets are currently used in state-of-the-art systems for a number of tasks in computer vision, and to date, the three most recent systems to achieve state-of-the-art performance in 3D object recognition on the ModelNet40 \citep{3DShapeNets} benchmark have made use of 2D ConvNets pre-trained on ImageNet \citep{MVCNN}\citep{Pairwise} \citep{FusionNets} and evaluated using multiple rendered object views.

Voxel models, wherein object shape is represented as a binary occupancy grid, provide a representation suitable for use with ConvNets, but present a number of difficulties. The addition of a third spatial dimension in the regular grid comes with a corresponding computational cost, and the curse of dimensionality is a central issue, limiting the available resolution of the voxel grid. Low resolution grids make it difficult to differentiate between similar shapes,and toss some of the texture information available in 2D renderings of equivalent dimesnionality. Shallow 3D ConvNets have been evaluated on the ModelNet benchmark \citep{VoxNet} \citep{ORION}, but are generally outperformed by multi-view 2D ConvNets.

Despite these challenges, we posit that deep ConvNets are viable for use in modeling voxel-based 3D objects for both generative and discriminative tasks. In this work, we present deep ConvNet architectures for both generative and discriminative voxel modeling, and explore issues specific to voxel-based representations. Our generative methods display high  fidelity shape interpolation, and our discriminative methods outperform the current state of the art by a relative 51.5\% and 53.2\% on the ModelNet40 and ModelNet10 benchmarks.

\begin{figure}[ht]
  \centering
  \includegraphics[scale=0.33]{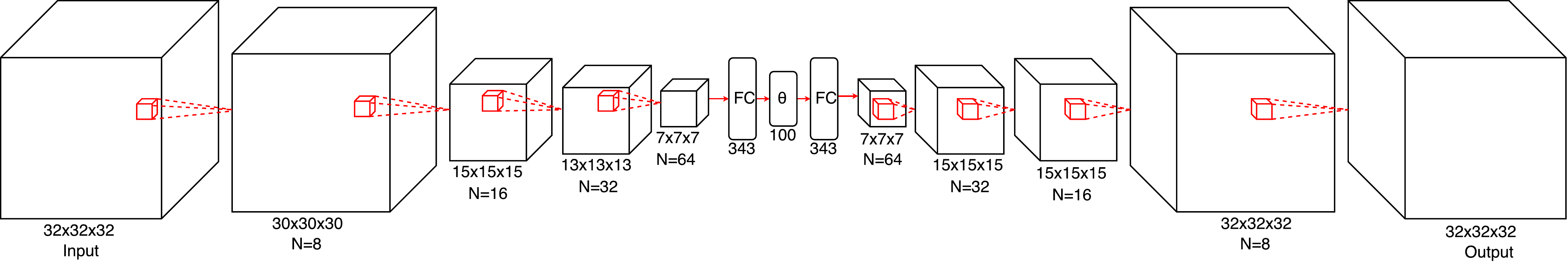}
  \caption{VAE Architecture.}
  \label{VAE_arch}
\end{figure}

\section{Voxel-Based Variational Autoencoders}
\label{VAE}
Interpolating between binary grids permits no obvious mathematical interpretation; if we wish to learn how voxellated objects relate to one another, we require a model capable of reasoning in an abstract feature space that captures the salient factors of variation. For this work, we select the Variational Autoencoder (VAE) \citep{VAE}, a probabilistic framework that learns both an inference network to map from an input space to a set of descriptive latent variables, and a generative network that maps from the latent space back to the input space. By training a network to infer the latent variables which describe the underlying factors of variation between objects, we gain the ability to smoothly transition between objects by interpolating between each object's latent description and reconstructing using the decoder network.
\subsection{Model Architecture}
Our model, implemented in Theano\citep{Theano} with Lasagne,\footnote{https://github.com/Lasagne/Lasagne} comprises an encoder network, the latent layer, and a decoder network, as displayed in Figure~\ref{VAE_arch}. The encoder network consists of 4 convolutional layers and a fully connected layer, followed by a linear projection from the fully connected layer to the latent layer. The decoder network has an identical, but inverted, architecture, and its weights are not tied to the encoder's. Each convolutional layer has a bank of 3x3x3 filters, starting with 8 filters in the layer furthest from the latents and doubling at each subsequent layer. 

All layers use the exponential linear unit \citep{ELU} nonlinearity, with the exception of the final layer, which uses a sigmoid nonlinearity. The output of each element of the final layer can be interpreted as the predicted probability that a voxel is present at a given location. 
Downsampling in the encoder network is accomplished via strided convolutions (as opposed to pooling) in every second layer. Upsampling in the decoder network is accomplished via fractionally strided convolutions, implemented as the gradient of an equivalent strided convolution\citep{arithmetic}, in every second layer.

The network is initialized with Glorot Initialization \citep{Glorot}, and all but the output layer are Batch Normalized\citep{Bnorm}. The variance and mean parameters of the latent layer are individually Batch Normalized, such that the output of the latent layer during training is still stochastic under the VAE parameterization trick. 

\subsection{Loss Function}
The loss function consists of the KL divergence prior on the latents, L2 weight regularization, and the reconstruction error, for which we use a specialized form of Binary Cross-Entropy (BCE).  The standard BCE loss is: 

\begin{center}$\mathcal{L} = -t\,log(o)\,-\,\,(1-t)\,log(1-o)$\end{center}
\begin{figure}[htbp]
  \centering
  \begin{tabular}{cc}
  \subf{\includegraphics[scale=.4]{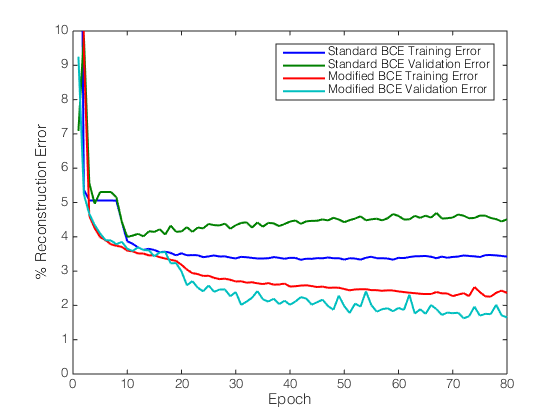}}{(a)}

&
 \subf{\includegraphics[scale=.25]{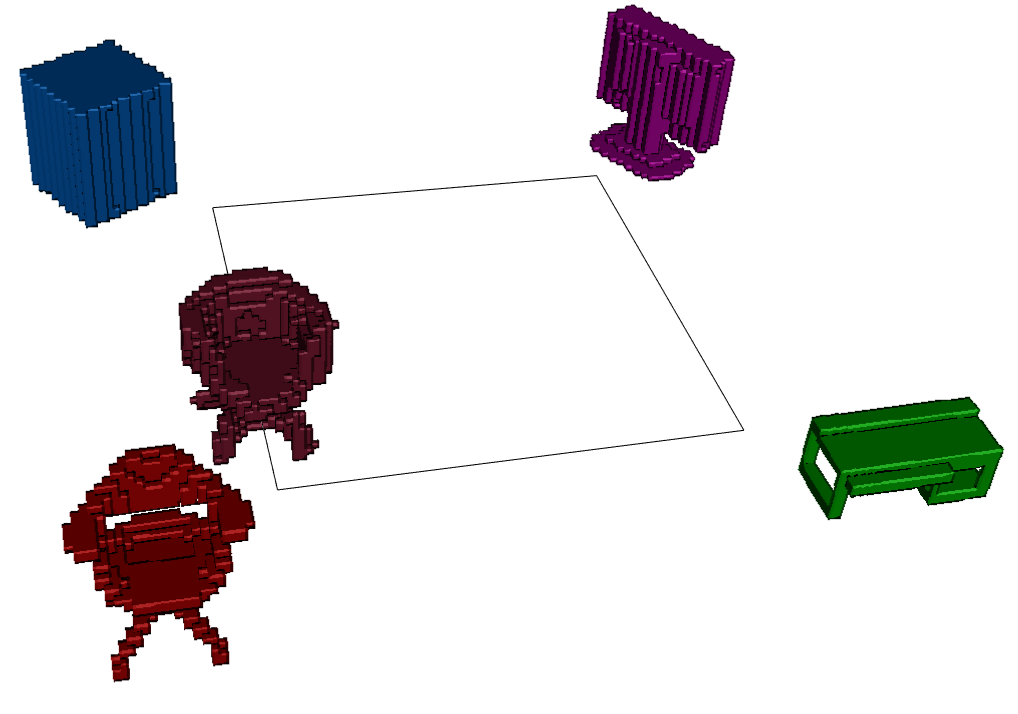}}{(b)}

 \end{tabular}
 \caption{Comparison of training regimes (a), user interface (b).}
 \label{VAE_training} 
\end{figure}


Where \textit{t} is the target value in \{0,1\} and \textit{o} is the output of the network in (0,1) at each output element. The derivative of the BCE with respect to \textit{o} severely diminishes as \textit{o} approaches \textit{t}, which can result in vanishing gradients during training. Additionally, the standard BCE weights false positives and false negatives equally; because over 95\% of the voxel grid in the training data is empty, the network can confidently plunge into a local optimum of the standard BCE by outputting all negatives.

We make two key modifications to the BCE to improve training. First, we change the range of the target and output to \{-1,2\} and [0.1,1),respectively. This change increases the magnitude of the loss gradient throughout the domain of \textit{o}, reducing the probability of vanishing gradients. Second, we add a hyperparameter $\gamma$ which weights the relative importance of false positives against false negatives:

\begin{center}$\mathcal{L} = -\gamma\,t\,log(o)\,-\,(1-\gamma)\,(1-t)\,log(1-o)$\end{center}

During training, we set $\gamma$ to 0.97, strongly penalizing false negatives while reducing the penalty for false positives. Setting $\gamma$ too high results in noisy reconstructions, while setting $\gamma$ too low results in reconstructions which neglect salient object details and structure.

\subsection{Training}

The model is trained using stochastic gradient descent with Nesterov momentum \citep{Nesterov} for 100 epochs, or until the reconstruction error on a held-out validation set bottoms out. The learning rate is set to 0.0001 for the first epoch, then increased to 0.001. The data is augmented by adding random translations and horizontal flips to each training example, as in \citep{VoxNet}, then training on one noisy and one uncorrupted copy of each instance, randomly shuffled. By training the network to reconstruct both corrupted and uncorrupted data, we force it to learn invariance to small structural variations.

We first validate our modification to the BCE by comparing the validation errors of two identically initialized networks, one trained with the standard BCE, and one trained with our modification. The reconstruction error is plotted against training epochs in Figure~\ref{VAE_training}(a). Interestingly, we note that the validation error is lower than the training error for this particular training run. We experiment with a number of different model architectures and training regimes before converging on the final method detailed above. In particular, we experiment with augmenting the training objective by adding a 10-unit fully connected softmax layer for classification in parallel with latent estimation, as well as a denoising objective, neither of which result in any observable performance improvement.

\subsection{User Interface}
We present a graphical user interface modeled after \citep{GUI_REF}. The interface, implemented using VTK\citep{VTK}, allows the user to drag a center object that interpolates between up to four different objects, and supports class-unconditional random shape generation. The interpolant endpoint models are randomly selected from the ModelNet10 test set at runtime, and both inference and reconstruction run in real time on a laptop with a GT730m graphics card. Figure~\ref{VAE_training}(b) shows a screenshot of the interface, and a video of the interface in action is available online.\footnote{https://www.youtube.com/watch?v=LtpU1yBStlU}  

\begin{figure}[htbp]
  \centering
  \begin{tabular}{ccc}
  \subf{\includegraphics[scale=.45]{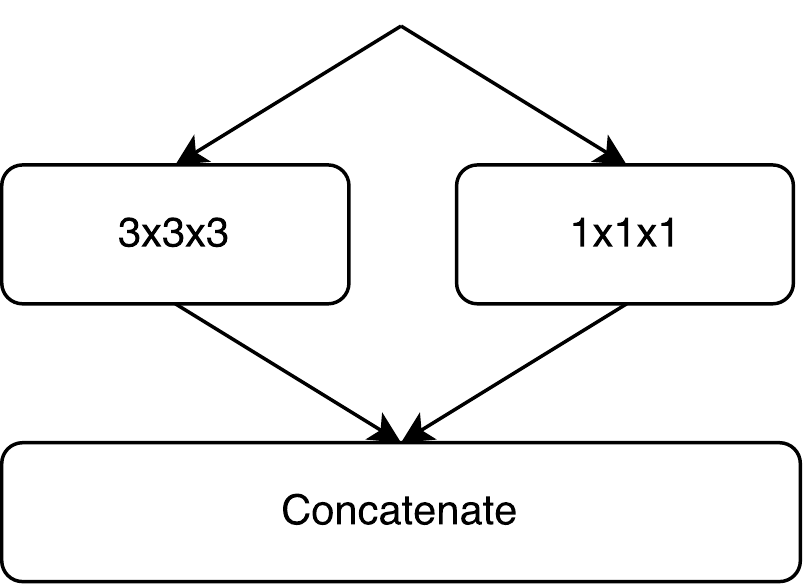}}
  {Voxception}

&
 \subf{\includegraphics[scale=.5]{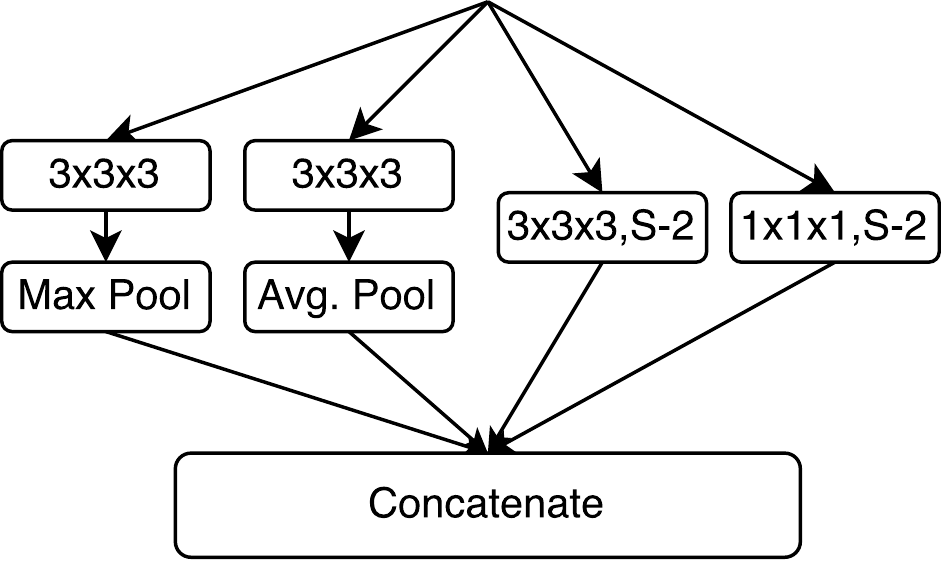}}
 {Voxception-Downsample}
&
\subf{\includegraphics[scale=.5]{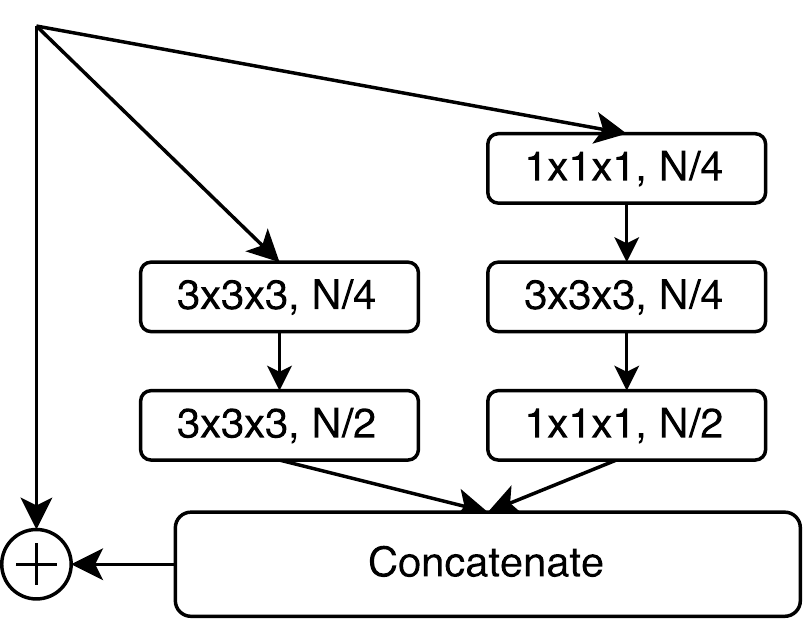}}
{Voxception-Resnet}  
 \end{tabular}
 \caption{Voxception and Voxception-Resnet Blocks.}
 \label{VRNBLOCKS} 
\end{figure}

\section{Voxel-Based Convnets for Classification}
\label{CLASSIFICATION}
Voxel-based ConvNets were first applied to 3D object recognition in Voxnet\citep{VoxNet}, a shallow volumetric ConvNet architecture, and were also used in ORION\citep{ORION}, wherein the classification task was augmented with an orientation estimation task. A recent extension, FusionNets\citep{FusionNets}, combines ConvNets trained on voxellated models and pre-trained ConvNets fine-tuned on rendered object views.

\subsection{Architecture}
Our model is designed in line with approaches used for high performance 2D ConvNets for object classification. Key to our approach is the use of Inception-style modules\citep{Inception}, Batch Normalization\citep{Bnorm}, Residual connections with pre-activation \citep{ResNet}\citep{PreActivation} and stochastic network depth \citep{StochasticDepth}. In contrast to previous 3D ConvNet approaches which used shallow networks, we train networks with up to 45 layers to take advantage of the increased expressivity that comes with model depth. Compared to FusionNets\citep{FusionNets}, our model requires significantly fewer parameters (18M as opposed to 118M for a full FusionNet) and fewer object views (12 or 24, compared to 60). Code to train and test our models is publicly available.\footnote{https://github.com/ajbrock/Generative-and-Discriminative-Voxel-Modeling}

\subsubsection{Voxception}
After initial tests with vanilla ConvNets, we adopted a simple Inception-style architecture. The intuition behind the design was to maximize the number of possible "pathways" for information to propagate through the network, while still maintaining simplicity and efficiency. 

For non-downsampling layers (Figure~\ref{VRNBLOCKS}, left) , we concatenate equal numbers of 1x1x1 and 3x3x3 filters, allowing the network to choose between taking a weighted average of the featuremaps in the previous layer (i.e. by heavily weighting the 1x1x1 convolutions) or focusing on spatial relationships (i.e. by heavily weighting the 3x3x3 filters). For downsampling layers (Figure~\ref{VRNBLOCKS}, center), we stack 3x3x3 convolutions with strided pooling operations, using both max and average pooling, and concatenate those features with strided 3x3x3 and 1x1x1 convolutions. Our intent is for the downsampling layers to let the network learn the best relative weighting of the various downsampling methods, maximizing propagation of information while still producing a more compact representation.

Our final model is nine layers deep, with four Voxception blocks and three Voxception Downsample blocks, followed by two fully connected layers and a softmax nonlinearity. 

\subsubsection{Voxception-ResNet}
The Voxception-ResNet (VRN) architecture is based on the ResNet architecture\citep{ResNet}, but concatenates both the ResNet Bottleneck Block and the standard ResNet block into a single Inception\citep{Inception}-style block (Figure~\ref{VRNBLOCKS}, right). To improve parameter efficiency, the early layers in each path of the block have half as many filters as the final layer. Downsampling is accomplished through Voxception-Downsample blocks, which we do not change in our ResNet model. We change the order of application of rectifying nonlinearities and Batch Normalization to obtain pre-activation blocks\citep{PreActivation}. Finally, we stochastically drop the non-residual paths of blocks\citep{StochasticDepth}, where the keep probability is linearly decreased from 1.0 in the first layer to 0.25 in the final VRN layer, and used as a weighting value instead of a drop probability at test time.

\begin{figure}[ht]
  \centering
  \includegraphics[scale=.6]{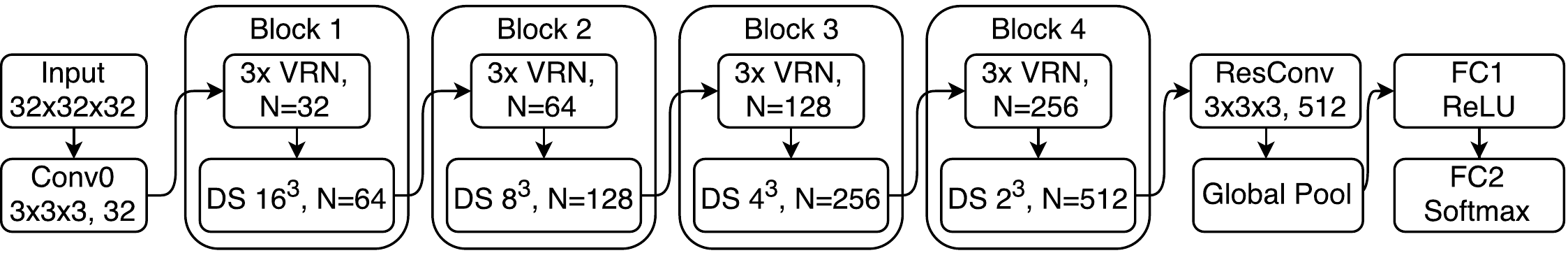}
  \caption{Voxception-ResNet 45 Layer Architecture. DS are Voxception-Downsample blocks.}
  \label{VRNFULL}
\end{figure}
Our best-performing architecture is shown in Figure~\ref{VRNFULL}, and consists of an initial convolutional layer, four main units, each containing three stacked VRN blocks and a Voxception-Downsample block, a final convolution with a residual connection and keep probability of 0.5, then a global pooling layer and two fully-connected layers. The number of filters begins at 32, and is doubled at each downsampling block. The deepest path through the network is 45 layers (going through the 3-layer section of the VRN block), and the shallowest path (assuming all droppable non-residual paths are dropped) is 8 layers deep.

\subsection{Data Augmentation and Training}

Unless otherwise specified, all models were initialized using Orthogonal Initialization\citep{orthog}, Batch Normalized\citep{Bnorm}, and trained using Nesterov Momentum\citep{Nesterov} with a momentum value of 0.9. Other than the final softmax nonlinearity, Exponential Linear Units\citep{ELU} are used as activations throughout the network. We change the binary voxel range from \{0,1\} to \{-1,5\} to encourage the network to pay more attention to positive entries.

We experiment with multiple learning rate decay schemes and find that dividing the learning rate by a factor of 2 every time the validation loss bottoms out to be more effective than annealing at a constant rate after a set number of minibatches or epochs. Initial hyperparameter studies were validated using a held-out, class-balanced tenth of the training set, but for final evaluation the annealing schedule was fixed and the entire training set was used. 

During each epoch, we train on two copies of each example, where one of the copies is randomly flipped about a horizontal axis and/or translated, as was done in VoxNet\citep{VoxNet}. We use a fixed random seed scheme to ensure that different training runs make use of identically augmented datasets. We use two versions of the training set, one with 12 rotations of each instance, and one with 24 rotations of each instance. For our best performing models, we warm up the network by training for twelve epochs on the 12-rotation training set, then anneal the learning rate and fine-tune on the 24-rotation training set. During testing, we measured predictions on a single view and averaged predictions across 24 rotated copies of each instance. We found this data augmentation to be essential for training deeper networks, especially Voxception-Resnet. We produce a simple ensemble by summing predictions from five VRN models and one Voxception model. A single training epoch, using a batch size of 50 and the full 24-rotation augmented dataset, takes around 6 hours on a single Titan X, and most models require around 6 days of training to converge.

We also experiment with treating the different rotations as separate channels for a single instance (analagous to RGB channels in a 2D image) but found that training a model to look at a single orientation and averaging predictions across rotated versions of an instance yielded better performance. We suspect that this is because, without otherwise changing the model architecture, the rotation-channel network must still pass information through equivalent representational bottlenecks, as opposed to being able to separately evaluate each rotation as individual instances. Additionally, we experimented with a variety of values for the range of the binary voxel grid, including an adaptive method wherein the numerical value of positive entries is equal to the 100 times the percentage of the grid occupied by the object, but did not find these methods to significantly alter performance.

We experimented with stacking 3x1x1,1x3x1,1x1x3 blocks in place of 3x3x3 convolutions, but found that this did not noticeably affect performance or training time. We tried to use the Adam and Adamax\citep{Adam} optimizers; although using these optimization schemes caused training error to bottom out very quickly, the validation error did not improve with the training error, suggesting that the model was rapidly overfitting to the training set.
 
\begin{figure}[h]
  	
  \centering
  \includegraphics[scale=.25]{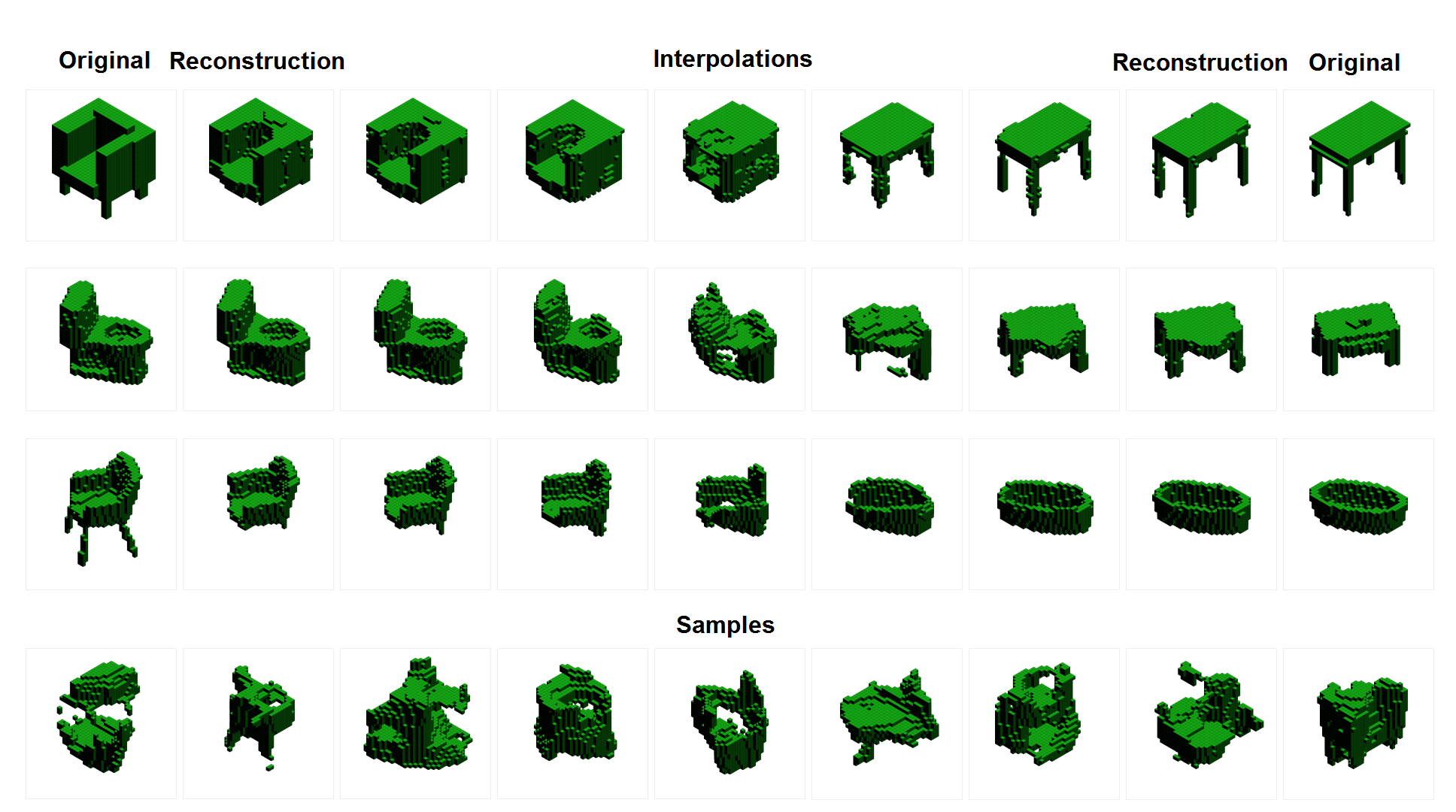}
  \caption{Voxel-Based VAE Reconstructions, Interpolations, and Samples.}
  \label{VAERECONfig}
\end{figure}

We also found that aggressively downsampling early in the network by replacing the initial convolution with a strided convolution enabled us to train significantly deeper models, but that such networks underperformed their 45 layer counterparts. We theorize that placing an early representational bottleneck causes the network to toss features that it might otherwise learn to keep or propagate to deeper layers, placing an upper bound on performance, or that we simply did not provide sufficient data to train networks of such depth.

\section{Results}
\label{RESULTS}

\subsection{Voxel-based VAE}

The reconstruction accuracy of our fully-trained VAE, evaluated on the ModelNet10 test set, is displayed in Table~\ref{reconstruction-table}. The model attains a 99.39\% true positive and 92.36\% true negative reconstruction accuracy on the ModelNet10 test set, indicating that it learns to reconstruct with high fidelity, but tends to slightly overestimate the probability of a voxel being present. Reconstruction and interpolation examples are displayed in Figure~\ref{VAERECONfig} alongside class-unconditional random samples.

\subsection{Object Classification}

The accuracy of our discriminative models is evaluated and compared against competing approaches in Table~\ref{Class-table}. Our best single VRN model obtains 91.33\% accuracy on the ModelNet40 test set, and 93.61\% accuracy on the ModelNet10 test set, which are respectively better than any previous published results and competitive with the current state of the art. Our best ensemble of VRN models obtains state-of-the art 95.54\% and 97.14\% accuracy on both ModelNet subsets, improving the state of the art by a relative 51.5\% and 53.2\%, respectively.  Our best VRN model obtains 88.98\% accuracy when tested using a single view of the object. Our best ensemble of models trained solely on ModelNet10 obtains 94.71\% accuracy, which is also better than any previously published result; we report our ModelNet10 accuracy for models trained on ModelNet40 for consistency with previous results\citep{MVCNN}\citep{Pairwise}\citep{FusionNets}.

\begin{table}[ht]
  \centering
  \setlength\tabcolsep{4pt}
  \begin{minipage}{.5\textwidth}
  \centering

  \caption{Reconstruction Results.}
  \label{reconstruction-table}
  \begin{tabular}{cccc}
  	
    \toprule
	
        & Predicted: & Predicted: \\ 
        & Positive     & Negative \\
    \midrule
    Actual: Positive & 99.39\%  & 0.61\%     \\
    Actual: Negative & 7.64\% & 92.36\%    \\
    \bottomrule
  \end{tabular}
  
\end{minipage}\begin{minipage}{.5\textwidth}
  \centering

  \caption{Classification Results.}
  \label{Class-table}
  \begin{tabular}{ccc}
    \toprule
    Model & ModelNet40 &  ModelNet10    \\
    \midrule
    
  	Voxnet & 83.00\% & 92.00\%\\
  	MVCNN & 90.10\% & - \\
  	Pairwise & 90.70\% & 92.80\%\\
  	FusionNets & 90.80\% & 93.11\%\\
  	ORION & - & 93.80\% \\
  	Voxception & 90.56\% & 93.28\%\\
  	VRN (One-View) & 88.98\% & - \\
  	VRN & 91.33\% & 93.61\%\\
  	VRN Ensemble & {\bf 95.54\%} & {\bf 97.14\%}\\
  
    \bottomrule

  \end{tabular}
\end{minipage}
\end{table}
\section{Discussion}

\subsection{Voxel-Based VAE}
The network achieves passable reconstruction accuracy, and learns to smoothly interpolate between arbitrary, previously unseen shapes. The network is additionally capable of generating random shapes with consistent structure, indicating that the learned latent space is successful in disentangling the factors of structural variation, though these new shapes.

The network performs well for dense objects, particularly thick dense objects such as sofas and toilets, but occasionally struggles to reconstruct objects with long, thin members, such as tables or chairs. We suspect these features are too small to activate in the receptive field of the appropriate latents, and are lost in favor of denser features which weigh more heavily in the loss function, and suggest imposing an additional "local" reconstruction function that measures reconstruction accuracy in subsets of the voxel grid, such that the network learns to reconstruct features regardless of how small they are relative to the entire object.

The network also struggles to reproduce crisp edges, preferring to output smooth, rounded edges. We posit that this is analogous to the way in which a vanilla 2D VAE will tend to output images with blurry edges rather than crisp edges to avoid overconfidently making incorrect predictions.

\subsubsection{Interpolation}
The system is capable of smoothly interpolating between reconstructions, indicating that it learns a representation which captures the underlying factors of structural variation. For example, when interpolating between two objects of the same class but slightly different orientation, the model will make only minimal changes in the output during interpolation, rather than completely deconstructing and reconstructing the output (i.e. passing through the origin of the latent space).

When interpolating between drastically different objects, the interface exhibits a “flowing water” effect, wherein preexisting voxels will appear to smoothly shift between shapes, rather than appearing at random, as can be seen in Figure~\ref{VAERECONfig}.

\subsubsection{Sampling}
Samples generated by our model are shown in the last row of Figure~\ref{VAERECONfig}. Our samples consistently bear a semblance of structure, with few to no free-floating voxels, suggesting that the decoder network has learned to maintain output voxel connectivity regardless of the latent configuration. The major limitation of the VAE is that its generated samples do not, however, resemble real objects. We hypothesize that training a deeper, more expressive model on the ModelNet40 dataset, and augmenting the latent vector with a class-conditional vector, would enable the generation of objects which clearly belong to a particular class, but leave such an investigation to future work.

\subsection{Object Classification}
Our VRN model takes advantage of the increased expressivity associated with its substantially increased depth to achieve significant improvements on the ModelNet classification benchmark. 

We found that averaging predictions across rotations was critical to achieving top performance, but that even taking predictions from a single view resulted in passable 88.98\% ModelNet40 accuracy, and even an ensemble of 2 models predicting on a single view achieves over 91\% accuracy. We also found that ensembling with a mix of predictions averaged over 12 and 24 rotations resulted in even higher test performance on ModelNet40 (95.78\%), but we suspect that this result is not general, and do not claim it with our main results.

Our best single model is competitive with, but does not outperform the previous state of the art on ModelNet10, ORION\citep{ORION}, which augments the classification task with a rotation estimation task. We suspect that our deeper models do not perform quite as well when trained on the smaller subset due to a lack of data, with only 3991 training instances in ModelNet10 compared to 9843 instances in ModelNet40. Additionally, ORION\citep{ORION} incorporates class-specific priors to determine precisely which rotations to train on, which we eschew in favor of generality.

Our (by no means novel) hypothesis that the upper bound on model depth is dependent on the amount of available data is consistent with our observations that significant data augmentation was required to train our 45-layer model. We also note that our highest performance on ModelNet10 came from models trained on ModelNet40, despite ModelNet40 not containing any additional ModelNet10 instances. This is interesting from a transfer learning perspective: by learning to distinguish between a wider variety of classes, the model learns to better discriminate between a given subset of classes.

\subsubsection{Suggestions for Future Work}

We believe that there still remains plenty of low-hanging fruit to be gained by investigating deep ConvNets for 3D object classification, and provide several suggestions for improvement:

\begin{itemize}

\item Refine the voxel grid beyond 32x32x32 to increase spatial resolution and improve the available detail. 

\item Change the voxel grid to be real-valued based on the percentage of space occupied by the instance at each grid element. This could also be used with the VAE to allow it to represent more complex shapes by fitting a corner-connected polyhedron with volume equal to the predicted occupancy percentage.

\item Experiment with more data augmentation, such as rotating about random axes (or just adding more rotations about the central axis), random crops, or random rescaling.

\item Try out different Voxception architectures, downsampling methods, and activation functions. Our experiments focused primarily on high-level network architecture, and there are likely more effective ways to compose a Voxception-ResNet block.

\end{itemize}

\section{Conclusion}
We presented a voxel-based Variational Autoencoder and a graphical user interface for exploring the latent space of 3D generative models, along with a voxel-based deep convolutional neural network for classification. Our methods take into account challenges specific to voxel representations, and demonstrate the viability of voxel representations in discriminative tasks by improving the state of the art on the ModelNet classification task by large margins.

\subsubsection*{Acknowledgments}

This research was made possible by grants and support from Renishaw plc and the Edinburgh Centre For Robotics. The work presented herein is also partially funded under the European H2020 Programme BEACONING project, Grant Agreement nr. 687676.



\bibliographystyle{unsrtnat}
\bibliography{NIPS_paper_2016}

\begin{thebibliography}{21}
\providecommand{\natexlab}[1]{#1}
\providecommand{\url}[1]{\texttt{#1}}
\expandafter\ifx\csname urlstyle\endcsname\relax
  \providecommand{\doi}[1]{doi: #1}\else
  \providecommand{\doi}{doi: \begingroup \urlstyle{rm}\Url}\fi

\bibitem[Wu et~al.()Wu, Khosla, Yu, Zhang, Tang, and Xiao]{3DShapeNets}
Z.~Wu, A.~Khosla, F.~Yu, L.~Zhang, X.~Tang, and J.~Xiao.
\newblock 3d shapenets: A deep representation for volumetric shapes.
\newblock In \emph{CVPR 2015}.

\bibitem[Su et~al.()Su, Maji, Kalogerakis, and E.Learned-Miller]{MVCNN}
H.~Su, S.~Maji, E.~Kalogerakis, and E.Learned-Miller.
\newblock Multi-view convolutional neural networks for 3d shape recognition.
\newblock In \emph{ICCV 2015}.

\bibitem[Johns et~al.()Johns, Leutenegger, and Davison]{Pairwise}
E.~Johns, S.~Leutenegger, and A.~J. Davison.
\newblock Pairwise decomposition of image sequences for active multi-view
  recognition.
\newblock In \emph{CVPR 2016}.

\bibitem[Hegde and Zadeh(2016)]{FusionNets}
V.~Hegde and R.~Zadeh.
\newblock Fusionnet: 3d object classification using multiple data
  representations.
\newblock arXiv Preprint arXiv: 1607.05695, 2016.

\bibitem[Maturana and Scherer()]{VoxNet}
D.~Maturana and S.~Scherer.
\newblock Voxnet: A 3d convolutional neural network for real-time object
  recognition.
\newblock In \emph{IROS 2015}.

\bibitem[Sedaghat et~al.(2016)Sedaghat, Zolfaghari, and Brox]{ORION}
N.~Sedaghat, M.~Zolfaghari, and T.~Brox.
\newblock Orientation-boosted voxel nets for 3d object recognition.
\newblock arXiv Preprint arXiv: 1604.03351, 2016.

\bibitem[Kingma and Welling()]{VAE}
D.P. Kingma and M.~Welling.
\newblock Auto-encoding variational bayes.
\newblock In \emph{ICLR 2014}.

\bibitem[Team(2016)]{Theano}
The Theano~Development Team.
\newblock Theano: A python framework for fast computation of mathematical
  expressions.
\newblock arXiv Preprint arXiv: 1605.02688, 2016.

\bibitem[Clevert et~al.()Clevert, Unterthiner, and Hochreiter]{ELU}
D-A. Clevert, T.~Unterthiner, and S.~Hochreiter.
\newblock Fast and accurate deep network learning by exponential linear units
  (elus).
\newblock In \emph{ICLR 2016}.

\bibitem[Dumoulin and Visin(2016)]{arithmetic}
V.~Dumoulin and F.~Visin.
\newblock A guide to convolution arithmetic for deep learning.
\newblock arXiv Preprint arXiv: 1603.07285, 2016.

\bibitem[Glorot and Bengio()]{Glorot}
X.~Glorot and Y.~Bengio.
\newblock Understanding the difficulty of training deep feedforward neural
  networks.
\newblock In \emph{AISTATS 2010}.

\bibitem[Ioffe and Szegedy()]{Bnorm}
S.~Ioffe and C.~Szegedy.
\newblock Batch normalization: Accelerating deep network training by reducing
  internal covariate shift.
\newblock In \emph{ICML 2015}.

\bibitem[Sutskever et~al.()Sutskever, Martens, Dahl, and Hinton]{Nesterov}
I.~Sutskever, J.~Martens, G.~Dahl, and G.~Hinton.
\newblock On the importance of initialization and momentum in deep learning.
\newblock In \emph{ICML 2013}.

\bibitem[Yumer et~al.()Yumer, Asente, Mech, and Kara]{GUI_REF}
M.~Yumer, P.~Asente, R.~Mech, and L.~Kara.
\newblock Procedural modeling using autoencoder networks.
\newblock In \emph{UIST 2015}.

\bibitem[Schroeder et~al.(2006)Schroeder, Martin, and Lorenson.]{VTK}
W.~Schroeder, K.~Martin, and B.~Lorenson.
\newblock \emph{The Visualization Toolkit,}.
\newblock Kitware, 4 edition, 2006.

\bibitem[Szegedy et~al.(2016)Szegedy, Ioffe, and Vanhoucke]{Inception}
C.~Szegedy, S.~Ioffe, and V.~Vanhoucke.
\newblock Inception-v4, inception-resnet and the impact of residual connections
  on learning.
\newblock arXiv Preprint arXiv: 1602.07261, 2016.

\bibitem[He et~al.()He, Zhang, Ren, and Sun]{ResNet}
K.~He, X.~Zhang, S.~Ren, and J.~Sun.
\newblock Deep residual learning for image recognition.
\newblock In \emph{CVPR 2016}.

\bibitem[He et~al.(2016)He, Zhang, Ren, and Sun]{PreActivation}
K.~He, X.~Zhang, S.~Ren, and J.~Sun.
\newblock Identity mappings in deep residual networks.
\newblock arXiv Preprint arXiv: 1603.05027, 2016.

\bibitem[Huang et~al.(2016)Huang, Sun, Liu, Sedra, and
  Weinberger]{StochasticDepth}
G.~Huang, Y.~Sun, Z.~Liu, D.~Sedra, and K.~Q. Weinberger.
\newblock Deep networks with stochastic depth.
\newblock arXiv Preprint arXiv: 1603.09382, 2016.

\bibitem[Saxe et~al.()Saxe, McClelland, and Ganguli]{orthog}
A.M. Saxe, J.~L. McClelland, and S.~Ganguli.
\newblock Exact solutions to the nonlinear dynamics of learning in deep linear
  neural networks.
\newblock In \emph{ICLR 2014}.

\bibitem[Kingma and Ba(2014)]{Adam}
D.P. Kingma and J.~Ba.
\newblock Adam: A method for stochastic optimization.
\newblock arXiv Preprint arXiv: 1412.6980, 2014.

\end{thebibliography}

\end{document}